\documentclass[runningheads]{llncs}
\usepackage{graphicx}
\usepackage[normalem]{ulem}
\usepackage{microtype}

\usepackage{times}
\usepackage{latexsym}
\usepackage[T1]{fontenc}
\usepackage[utf8]{inputenc}
\usepackage{microtype}
\usepackage{multirow}
\usepackage{multicol}

\usepackage{inconsolata}
\usepackage{graphicx}
\usepackage{amsmath,amsfonts}
\usepackage{bm}
\usepackage{soul}
\usepackage{xcolor}
\usepackage{color}
\usepackage{booktabs} 
\usepackage{dsfont}
\usepackage{subfigure}
\usepackage[normalem]{ulem}

\begin{document}
\title{Large Language Models are Diverse Role-Players \\ for Summarization Evaluation}
%
%
%
%
%

\author{Ning Wu \and
Ming Gong\thanks{Corresponding author.} \and
Linjun Shou \and
Shining Liang \and
Daxin Jiang} 
\authorrunning{N. Wu et al.}
\institute{STCA Search \& Distribution Group, Microsoft, China\\
\email{wuning,migon,lisho,shiningliang,djiang@microsoft.com}}
%

\maketitle              
\begin{abstract}
Text summarization has a wide range of applications in many scenarios. The evaluation of the quality of the generated text is a complex problem. A big challenge to language evaluation is that there is a clear divergence between existing metrics and human evaluation. A document summary’s quality can be assessed by human annotators on various criteria, both objective ones like grammar and correctness, and subjective ones like informativeness, succinctness, and appeal.
Most of the automatic evaluation methods like BLUE/ROUGE may be not able to adequately capture the above dimensions. In this paper, we propose a new evaluation framework based on LLMs, which provides a comprehensive evaluation framework by comparing generated text and reference text from both objective and subjective aspects. First, we propose to model objective and subjective dimensions of generated text based on roleplayers prompting mechanism. Furthermore, we introduce a context-based prompting mechanism that is able to generate dynamic roleplayer profiles based on input context. Finally, we design a multi-roleplayer prompting technology based on batch prompting and integrate multiple outputs into the final evaluation results.   Experimental results on three real datasets for summarization show that our model is highly competitive and has a very high consistency with human annotators.

\end{abstract}
\section{Introduction} \label{sec:intro}
Text summarization has wide applications in various research and application fields.  Recently, some works found that there is a clear gap between existed metrics like BLEU~\cite{papineni2002bleu}, ROUGE, BertScore~\cite{zhang2019bertscore} and human annotations~\cite{goyal2022news,yuan2022selecting}.
Although typical overlap-based and model-based metrics can capture lexicon level or semantic level similarity between generated text and reference text, specific dimensions like coherence, grammar, and interestingness still can't be captured. As depicted in Figure 1, the summarization task reveals the inadequacy of traditional metrics such as BLUE/ROUGE: they are unable to reflect the true quality of the text after reaching a certain level. To achieve consistency between human evaluation and automatic metrics, we encounter two main challenges: 1) How to model objective criteria of evaluation such as coherence and grammar. 2) How to model subjective criteria of evaluation such as interestingness~\cite{gao2014modeling,hidi1986interestingness}, comprehensiveness, and usefulness from the standpoint of users. Natural language has various modes of expression for the same concept, so assessing its quality based on a few static criteria is hard.

\begin{figure}[!ht]
    \centering
    \includegraphics[width=0.8\linewidth]{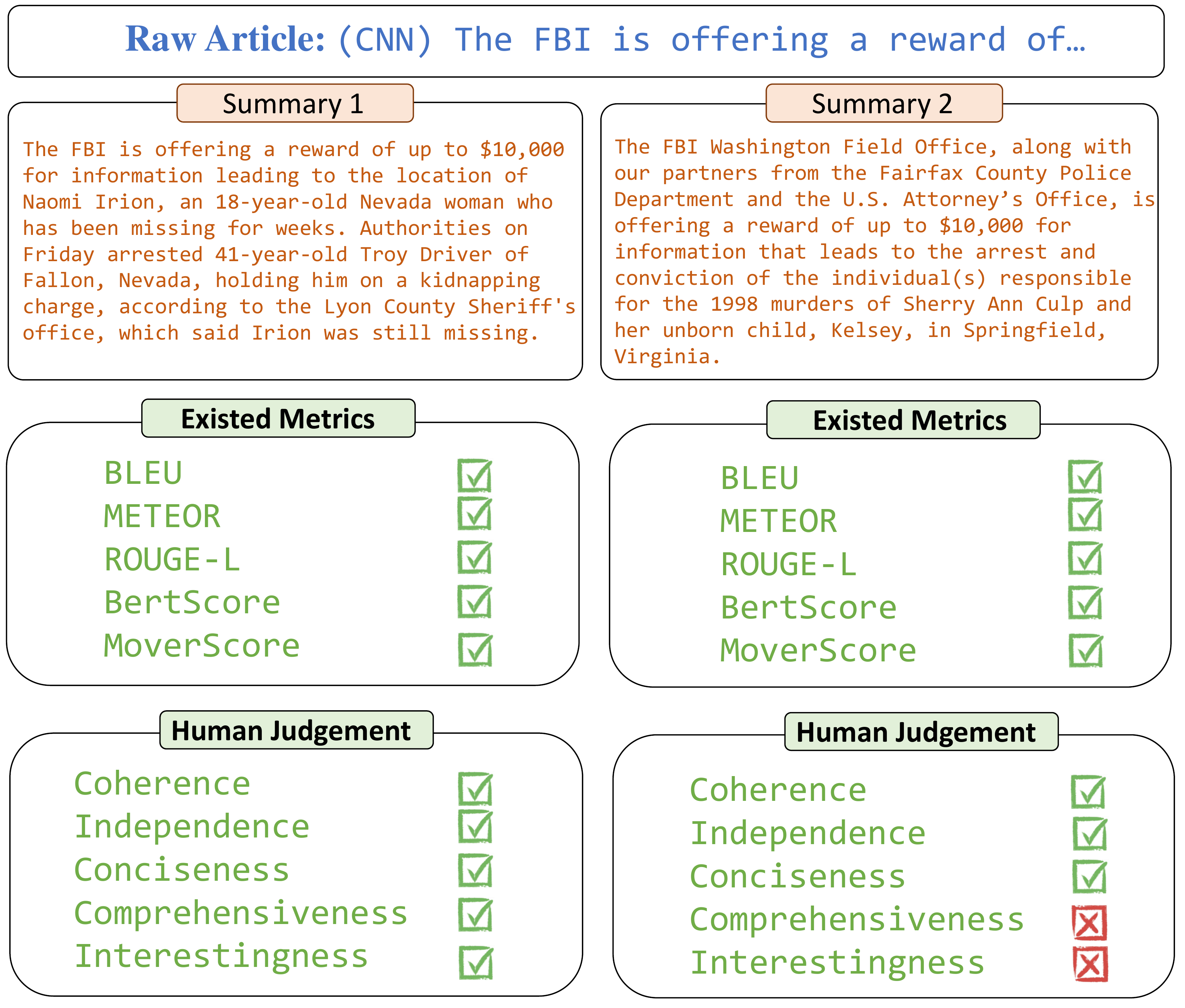}
    \caption{Two summarizations of CNN News, they are generated by two models (GPT3~\cite{brown2020language}, T0~\cite{sanh2021multitask}), and have similar BLEU and ROUGE metrics, but the second summary is obviously worse than the first one on two more complicated dimensions. }
    \label{first}
\end{figure}

Motivated by the ability of LLMs to handle multi-domains, we investigate how to leverage LLMs for measurement in this paper. Since it is difficult to make LLMs provide a consistent and fair score for the generated text~\cite{wang2022self}, we propose a comparison-based evaluation method to quantify the quality of the generated text, namely DRPE, which stands for \uline{D}iverse \uline{R}ole-\uline{P}layer for Summarization Generation \uline{E}valuation. In particular, we devise a roleplayer-based prompting strategy in this system for objective and subjective dimension measurement. Our method comprises two parts: 1) Static roles construction and dynamic roles prompts generation. 2) A multi-roleplayer framework to conduct a comprehensive evaluation. 

For a given generation task, its measurement can be broken down into several dimensions. Typical objective metrics such as coherence and grammar are relatively easy to be agreed upon by most people, so we manually created static roles for each objective dimension of the task. It is expressed as <\emph{Judger types}, \emph{Judger description}>. With a static role, we prompt LLM by asking it to impersonate a real judger based on the judger type and description and then vote for the better option. Furthermore, a comprehensive measurement is usually complex and dynamic. Depending on different cases in a summarization task, different aspects need to be taken into account. Therefore, we propose to dynamically generate some potential users based on the content and let LLMs conduct subjective measurements on behalf of these users. The dynamic roles can be expressed as <\emph{User types}, \emph{User description}>. Lastly, we design a multi-roleplayer framework to eliminate redundant roleplayers and integrate the vote results of multiple roleplayers. Moreover, the multi-roleplayer framework can also enhance the stability of the LLM-based measurement system with relatively low inference costs. Experimental results show that our method significantly surpasses zero-shot LLMs and existing metrics on three typical summarization datasets with human annotations.

\section{Related Works}
\subsection{Large Language Model}
Large language model has been found to be capable of few-shot learning~\cite{brown2020language}. Chain-of-Thought~\cite{wei2022chain} is proposed to empower model reasoning capability for complex tasks. ZeroShot-Cot~\cite{kojima2022large} still shows relatively strong reasoning ability without any examples. Least-to-Most~\cite{zhou2022least} LLM decomposes a complex question into several sub-questions and solves these sub-questions in sequence and finally gives a complete answer to the original question. Recent work~\cite{goyal2022news} discovered that both reference-based
and reference-free automatic metrics cannot reliably evaluate zero-shot summaries. In this paper, we mainly explore the capability of LLM to compare the generated text and reference text.

\subsection{Existed Metrics}

The most widely used metric in machine translation is BLEU~\cite{papineni2002bleu}, which includes several modifications to Exact-$P_n$. A smoothed variant, SENTBLEU is computed at the sentence level. METEOR~\cite{banerjee2005meteor} computes Exact-$P_1$ and Exact-$R_1$ while allowing backing-off from exact unigram matching to matching word stems, synonyms, and paraphrases. ROUGE~\cite{lin2004rouge} is a commonly used metric for summarization evaluation. ROUGE-n computes Exact-$R_n$ (usually n = 1, 2), while ROUGE-L is a variant of Exact-$R_1$ with the numerator replaced by the length of the longest
common subsequence. BERTScore~\cite{zhang2019bertscore} computes a similarity score for each token in the candidate sentence with each token in the reference sentence. 
MoverScore~\cite{zhao2019moverscore} investigates the effectiveness of existing contextualized representations and Earth Mover’s Distance~\cite{rubner2000earth} for comparing system predictions and reference texts, leading to a new automated evaluation metric that achieves high correlation with human judgments of text quality.

\begin{figure*}[!ht]
    \centering
    \includegraphics[width=0.9\linewidth]{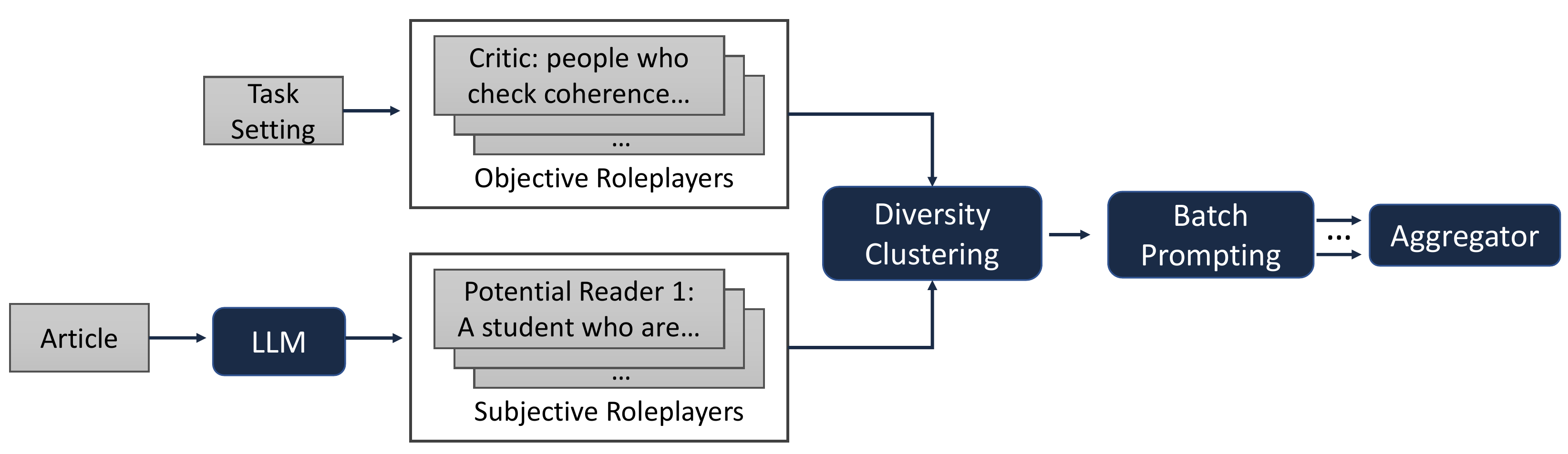}
    \caption{The overall framework of DRPE. Objective roleplayers are curated manually based on task setting, and subjective roleplayers are dynamically generated by LLMs. After diversity clustering, similar roles are eliminated, and all roles are played by LLMs to compare two candidates in batch prompting. Finally, results from multiple roles are aggregated. }
    \label{first}
\end{figure*}

\section{Methodology}
As discussed in Section~\ref{sec:intro}, currently, it forms a gap with human evaluation that automatic metrics for text generation stop at surface similarity (lexicon level or semantic level) which leads to biased perception and evaluation of the text generation capability of LLMs. In this section, we elaborate our proposed measurement framework for text generation primarily includes diversified roleplayers generation and roleplayers-based evaluation.

\subsection{Diversified RolePlayers Generation}~\label{sec:role_gen}
To build a novel framework differing from existing calculation-based automatic metrics, we decompose this task into objective and subjective dimensions and propose an LLM-based measurement method. In our framework, the LLMs act as a judge with a distinctive role to evaluate the text quality in a corresponding dimension and generate its evaluation results. Thus, we need to generate diversified roleplayers for objective and subjective dimensions at first.

\begin{figure}[!ht]
    \centering
    \includegraphics[width=0.9\linewidth]{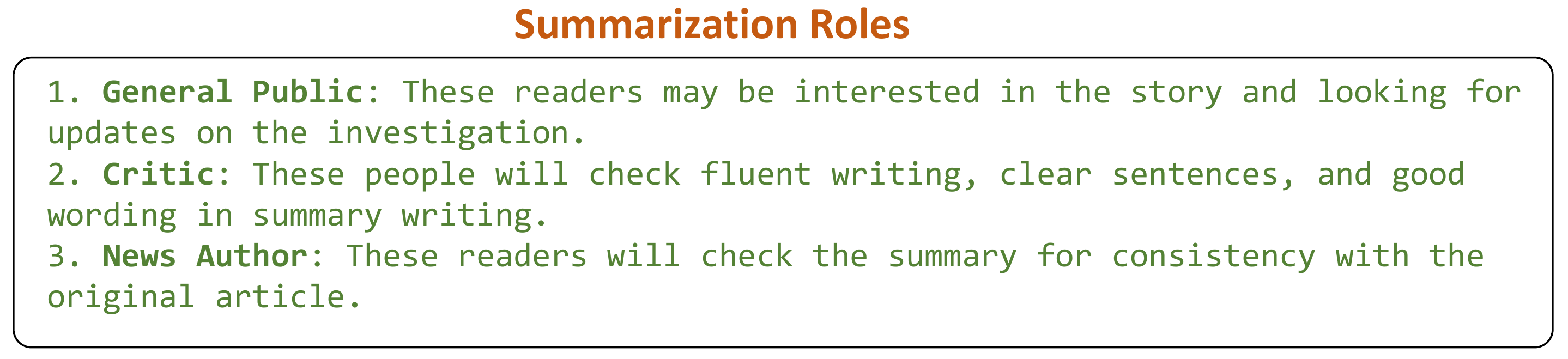}
    \caption{Three different static roles for summarization task. For different generation tasks, different aspects need to be taken into consideration. A good summary usually requires fluent writing, proper wording and capturing key points of raw article. }
    \label{fig-static}
\end{figure}


\subsubsection{Objective RolePlayers}
Overlap-based metrics such as BLUE/ROUGE measure lexicon-level consistency with n-grams. Model-based metrics like BERTScore can capture subtle syntax or semantic changes with embeddings similarity, but they have a limitation in evaluating high-quality text from LLMs. The parameter scale of BERT is much smaller than that of LLMs, so it cannot represent LLMs’ rich semantics.
Consequently in this paper, we take some advanced quality dimensions like fluency, consistency, grammar and coherence into consideration, which were rarely adopted before as they are difficult to be measured accurately. Since different tasks usually require different objective dimensions, and these dimensions are relatively easy to be agreed on by most people, hence we manually curated static objective dimensions for the summarization task and  make sure all these dimensions are fundamental and objective.  The static objective roles schema is presented below:<\emph{Judger type}, \emph{Judger description}>. where each \emph{Judger} works on one or multiple specific objective dimensions and \emph{Judger description} breaks down and exposit what specifically the Judger would focus on when evaluating. As shown in Figure~\ref{fig-static}, three different objective roles are designed for summarization tasks.

\begin{figure*}[!ht]
    \centering
    \includegraphics[width=0.9\linewidth]{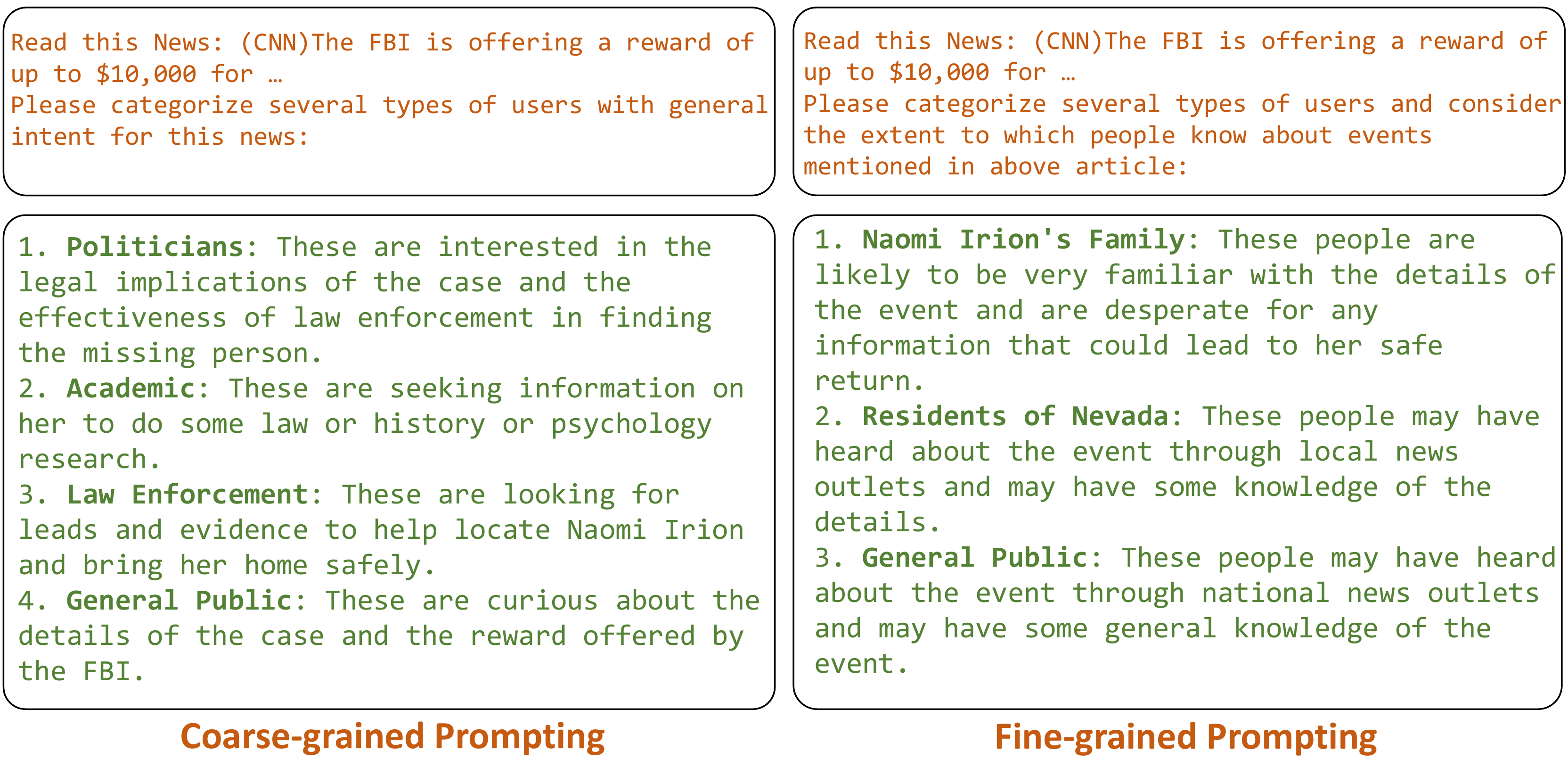}
    \caption{Coarse-grained and fine-grained prompting mechanism for comprehensive user profiles generation.  }
    \label{fig-dynamic}
\end{figure*}

\subsubsection{Subjective RolePlayers}
Text generation, unlike understanding tasks, does not have a perfect golden standard. Human-written material can only offer a high-quality example. Different readers may judge the text according to their own perspectives. For instance, consider the sports news about a renowned athlete in a game,
\begin{itemize}
    \item For the writing style, ordinary readers expect that it's concise and easy to understand, while journalists would pay attention to its structure and choices of words.
    \item For the content, causal fans like comprehensive data of the sports player and horizontal comparison with others, while die-hard fans are more eager for in-depth analysis of the sports player through data.
\end{itemize}

Therefore, we propose to collect subjective evaluations of model-generated text from diverse user perspectives, including whether they think the text is interesting, useful, etc. These dimensions are more abstract and difficult to quantify than objective dimensions which few studies have touched on and addressed to our knowledge. Specifically, we take each generated text as the context and prompt the LLM to generate its potential readers dynamically following the below schema: <\emph{User type}, \emph{User description}>. Here we design two user role generation prompts. As shown in Figure~\ref{fig-dynamic}, the former requires the LLM to consider the most common occupations with most people in the world which is coarse-grained and the latter aims to categorize people based on their familiarity with the text topics which is fine-grained. We merge objective judgers and subjective users generated by two kinds of prompting mechanisms as multi-role players for the next process. Considering that there may exist duplicate or similar users, we propose to conduct diversity clustering to improve measurement performance and reduce inference costs. First, each roleplayer type and its description are concatenated as the input of Sentence-BERT~\cite{reimers2019sentence} to obtain the representation. Next, we use the $k$-means algorithm to cluster roleplayers, and those closest to each cluster center are kept. Finally, the chosen role players will be leveraged for text evaluation.

\subsection{RolePlayer-based Evaluation}~\label{sec:txt_eval}
To mitigate the discrepancy between the human evaluation and automatic metrics, we propose to leverage the roleplayers as crowdsourcing voters to compare the summaries from multi-dimensions. Besides the static roleplayers that scrutinize the objective dimensions including grammar, fluency, coherence, etc, the dynamic roleplayers are generated according to the current article and simulate the psychology and behavior of the article readers (roles) to convey their subjective feelings. It's expected that our method could achieve higher consistency with human evaluation than existing automatic metrics focusing on surface similarity.

\subsubsection{Evaluation of RolePlayer}
\begin{figure*}[!ht]
    \centering
    \includegraphics[width=\linewidth]{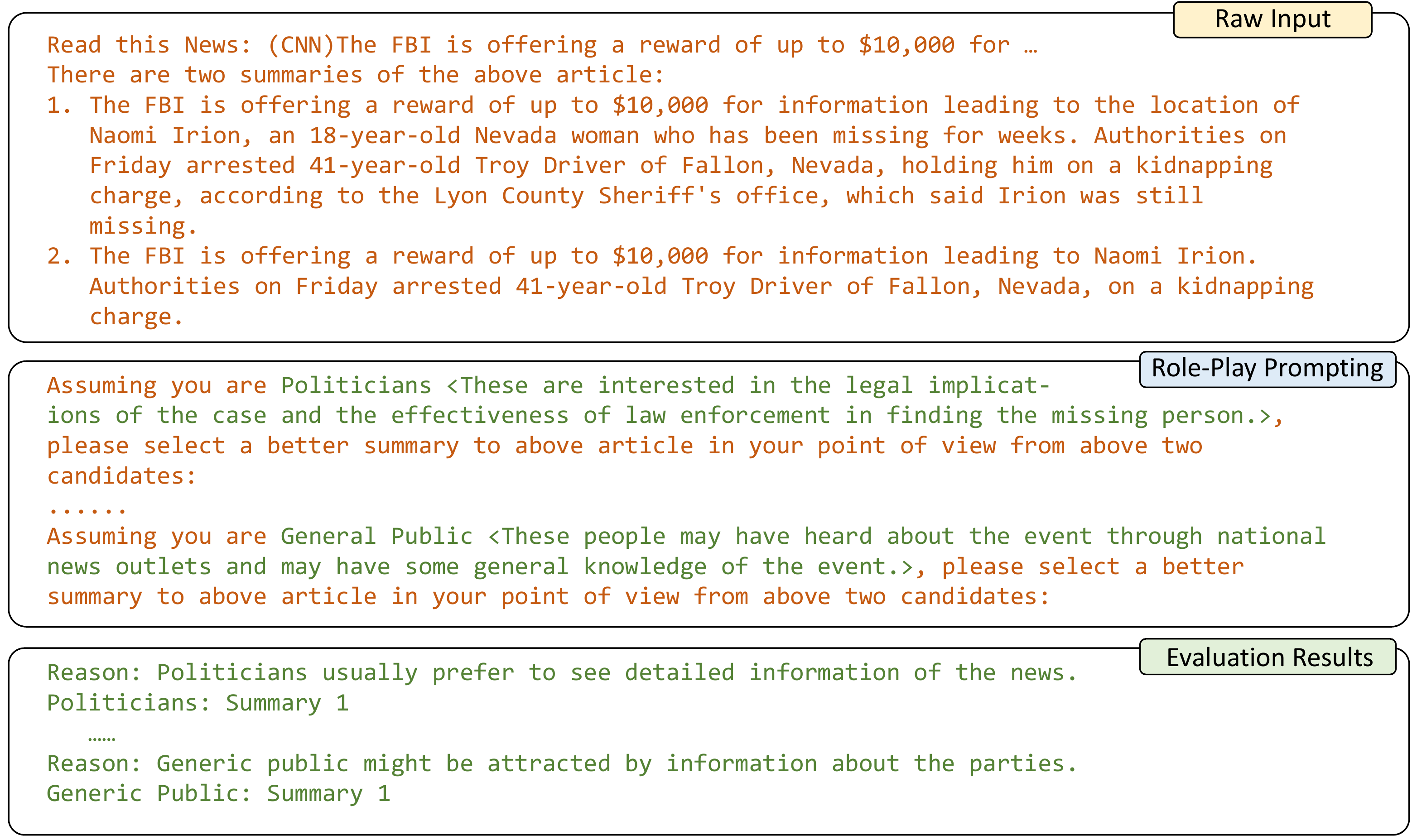}
    \caption{Compare generated summary and reference summary by multiple roleplayers with batch prompting. }
    \label{fig-batch}
\end{figure*}

Given the article $A$, we posit the reference summary generated by humans as $S$, and the candidate summary generated by models as $\hat{S}$, respectively. To evaluate, all the roleplayers perform pair-wise comparison as Figure~\ref{fig-batch}, since point-wise comparisons are inconsistent across different samples and list-wise comparisons are not stable. By our prompting, the LLMs play a specific role and output its analysis of two summaries, finally voting which summary is of better quality. In more detail, we parse and quantify the comparison result as $\hat{a}$:
\begin{equation}
    \hat{a} =
    \left\{ \begin{aligned}
        & 1    && {\text{If voting is candidate summary $\hat{S}$ }} \\
        & 0    &&  {\text{If voting is reference summary $S$ }} \\
    \end{aligned}
    \right.
\end{equation}

Assuming the DRPE score for generated text $\hat{S}$ is $\text{DRPE}(\hat{S}|A, S)$, it could be obtained by modeling the joint probability of reason and voting result as below:
\begin{align}\label{eq:DRPEScore}
    \text{DRPE}(\hat{S}|A, S) = \mathds{1}(\hat{a}=1)P(\hat{a}, \bm{r}|\bm{p}, A, S, \hat{S}, R)
\end{align}
where $\bm{r}$ is the comparison reason, $\bm{p}$ represents the prompt used here, $\hat{a}$ is voting result from LLMs, $A$ is raw article and $R$ is the role. To compute $P(\hat{a}, \bm{a})$, similar to~\cite{wang2022self}, we leverage the confidence when roleplayer generates its voting normalized by the output length~\cite{brown2020language},
\begin{align}
    & P(\hat{a}, \bm{r}|\bm{p}, A, S, \hat{S}, R) \nonumber \\
    & = \exp^{\frac{1}{K}\sum_{k=1}^K\log P(t_k|\bm{p}, A, S, \hat{S}, R, t_1, \dots, t_{k-1})}
\end{align}
where $\log P(t_k|\bm{p}, A, S, \hat{S}, R, t_1, \dots, t_{k-1})$ is the log probability of $k$-th token $t_k$ in $\bm{r}$ and $K$ is number of tokens.


\subsubsection{Batch Prompting}
To efficiently get comparison results for the summary pair ($S$, $\hat{S}$) from multiple roleplayers, we design a multi-roleplayer framework based on batch prompting to measure by both objective and subjective metrics in one-off inference. The different metrics are reflected by the objective and subjective roleplayers generated and clustered in Section~\ref{sec:role_gen}. 
As shown in Figure~\ref{fig-batch}, first, all the roleplayers are prompted to give votes for ($S$, $\hat{S}$) with respect to $A$ in their point of view, i.e., which summary could better describe the article, and tell us the reasons.
Then we aggregate the results to parse $\hat{a}=\{\hat{a}_j\}_{j=1}^N$ where $N$ is the number of roleplayers. According to Equation~\ref{eq:DRPEScore}, the multi-roleplayer DRPE score by batch prompting can be formulated as below:
\begin{align}
     & \text{DRPE}(\hat{S}|A, S) = P(\hat{a}|\bm{p}, A, S, \hat{S}) \nonumber \\
     & = \sum_{j=1}^N\mathds{1}(\hat{a}_j=1)P(\hat{a}_j, \bm{r}|\bm{p}, A, S, \hat{S}, R_j)
\end{align}
where $R_j$ denotes $j$-th roles. Compared with Self-Consistency CoT~\cite{wang2022self}, our framework decouples the answer (comparison result) and reasoning path (comparison reason), and brings higher inference efficiency as Self-Consistency CoT needs to generate each candidating answer separately while our method generates all voting results with once inference.


\section{Experiments}

\subsection{Experiments Setting}
\subsubsection{Datasets.}

 \textbullet \textbf{CNN2022}~\cite{nallapati2016abstractive,hermann2015teaching}: contains reference summaries that are approximately 3-4 sentences long. Summaries in this dataset are highly extractive and lead-biased. We use human study data on 100 recent articles from CNN, collected between March 1, 2022, and June 31, 2022. Each article is annotated by at least three judgers, and they are asked to choose the best and worst summaries from three candidates. To reduce noises, we only keep the best summary with at least 2 out of 3 votes and worst summary with at least 2 out of 3 votes. Finally, we obtain 81 best and worst summaries as candidate summaries with a piece of corresponding news. Finally, we use GPT-3 of the text-DaVinci-003 to generate reference summarization, and finally use our method to compare the candidate summary.

  \textbullet \textbf{BBC2022}~\cite{narayan2018don}: contains 1 sentence summaries of BBC news articles. In this dataset, references summaries, and consequently generated summaries from fine-tuned models are highly abstractive. We use human study data on 100 recent articles from BBC, collected between March 1, 2022, and June 31, 2022. We also take a similar preprocessing strategy like CNN2022 on this dataset.

  \textbullet \textbf{SummEval}~\cite{fabbri2021summeval}: contains a large and diverse collection of human judgments of model-generated summaries on the CNN/Daily Mail dataset annotated by both expert judges and crowd-source workers. For each news, we select two worst summaries and two best summaries according to their average scores on four dimensions~(coherence, consistency, fluency and relevance) labeled by experts. Finally, regarding one hundred news and corresponding reference summary in SummEval, we obtain 400 candidate summaries.

\subsubsection{Metrics.}

To measure the consistency between various metrics and human annotations, we follow the WMT18~\cite{ma2018results} standard practice and use absolute Pearson correlation $|\rho|$ to evaluate metric quality.


\subsubsection{Baselines.} 
Automatic metrics proposed for summarization evaluation can be broadly divided into two categories: (1) overlap-based metrics, specifically ROUGE METEOR and BLEU, and (2) similarity-based metrics that compute the similarity between embeddings representations of generated and reference summaries. Specifically, we report BERTScore and MoverScore. For LLMScore, we carefully design prompts for LLM, and directly use it to predict a better passage. 



 
 
 
 
 

\subsubsection{Implementation.}
We use the public GPT-3~\cite{brown2020language} of the text-DaVinci-003 version with 175B parameters from OpenAI for the LLMs implementation and use greedy decoding for inference with the temperature set to 0. We select this LLM because it has relatively good capability among public LLMs. Especially, we use three roles \textit{General Public}, \textit{Critic}, and \textit{News Author} which are described in Figure~\ref{fig-static}  as objective roleplayers in our DRPE, and we prompt the model to generate 4 dynamic roles for each case.

\subsection{Results}


\begin{table}[!ht]
  \centering
  \caption{Pearson correlation between several automatic metrics and human annotation. We bold the highest numbers for each dataset, and use AVG to denote the average scores on the three datasets. Results of GPT-D3 and DRPE are averaged over five runs with slight changing in prompts.}
  \setlength{\tabcolsep}{3.0mm}{
  \resizebox{0.8\columnwidth}{!}{%
    \begin{tabular}{l|ccccc}
    \toprule
\hspace{0.5em} Type &    Method      & CNN2022  & SummEval  & BBC2022 & AVG \\
\midrule
\multirow{5}{*}{\hspace{0.2em} Overlap} &    ROUGE-1  & 0.466  & 0.431 & 0.469  & 0.461  \\
&    ROUGE-2  & 0.437  & 0.354  & 0.443  & 0.411  \\
&    ROUGE-L  & 0.422  & 0.322  & 0.436  & 0.393  \\
&    BLEU & 0.475  & 0.372  & 0.502  & 0.450  \\
&    METEOR & 0.514  & 0.473  & 0.561  & 0.516  \\
\midrule
\multirow{2}{*}{Similarity} &      BERTScore & 0.554  & 0.455  & 0.568  & 0.526  \\
&    MoverScore & 0.456  & 0.385  & 0.442  & 0.428 \\
\midrule
\multirow{2}{*}{\quad LLM} &      GPT-D3 & 0.713  & 0.503  & 0.692  & 0.636  \\
&    DRPE & \textbf{0.816}  & \textbf{0.683}  & \textbf{0.784}  & \textbf{0.761}  \\

    \bottomrule
    \end{tabular}}%
    }
  \label{tab:main}%
\end{table}%

\subsubsection{Comparison with Existed Metrics}
Tables~\ref{tab:main} shows Pearson correlation to human judgments. We observe that typical overlap-based metrics generally performed badly and relevance-based metrics also underperformed. The simplest LLM-based method has a consistently better performance than BERTScore. Two types of LLM-based methods, GPT-D3, and DRPEScore have a clear gap between the other two methods. Especially, DRPEScore consistently performs better than GPT-D3.

\begin{table}[!ht]
  \centering
  \caption{Pearson correlation between several automatic metrics and human annotation. AVG denotes the average scores on the three datasets.}
  \setlength{\tabcolsep}{3.0mm}{
  \resizebox{0.8\columnwidth}{!}{%
    \begin{tabular}{l|cccc}
    \toprule
   Method      & CNN2022  & SummEval  & BBC2022 & Avg \\
\midrule
   DRPE & 0.816  & 0.683 & 0.784  & 0.761  \\
   w/o Batch Inferring  & 0.822  & 0.672  & 0.766  & 0.753  \\
   w/o Clustering  & 0.782  & 0.665  & 0.751  & 0.733  \\
   w/o Dynamic Roles  & 0.742  & 0.669  & 0.703  & 0.705  \\
   w/o Static Roles  & 0.734  & 0.604  & 0.711  & 0.683  \\

    \bottomrule
    \end{tabular}}%
    }
  \label{tab:abla}%
\end{table}%

\subsubsection{Ablation Study}
We conducted an ablation study on DRPE to gain insights into the detailed method design. We prepare four variants of our method: 
(1) \uline{w/o Batch Inferring} denotes without the batch prompting, each role is inferred alone;
(2) \uline{w/o Clustering} denotes without clustering mechansim;
(3) \uline{w/o Dynamic Roles} denotes without dynamic roles generation;
(4) \uline{w/o Static Roles} denotes without the human designed static roles.
Table~\ref{tab:abla} presents all comparison results of the four variants. As we can see, the performance rank on three datasets can be given as: w/o Static Roles < w/o Dynamic Roles < w/o Clustering < w/o Batch Inferring < DRPE. These results indicate that all components are essential to improve performance. And we can also find that batch inferring is able to save lots of inference tokens without losing performance. 


\begin{figure}[!ht]
\centering
\subfigure[Roles number k w.r.t correlation on CNN2022 dataset.]{
\begin{minipage}[t]{0.4\linewidth}
\centering
\includegraphics[width=0.95\linewidth]{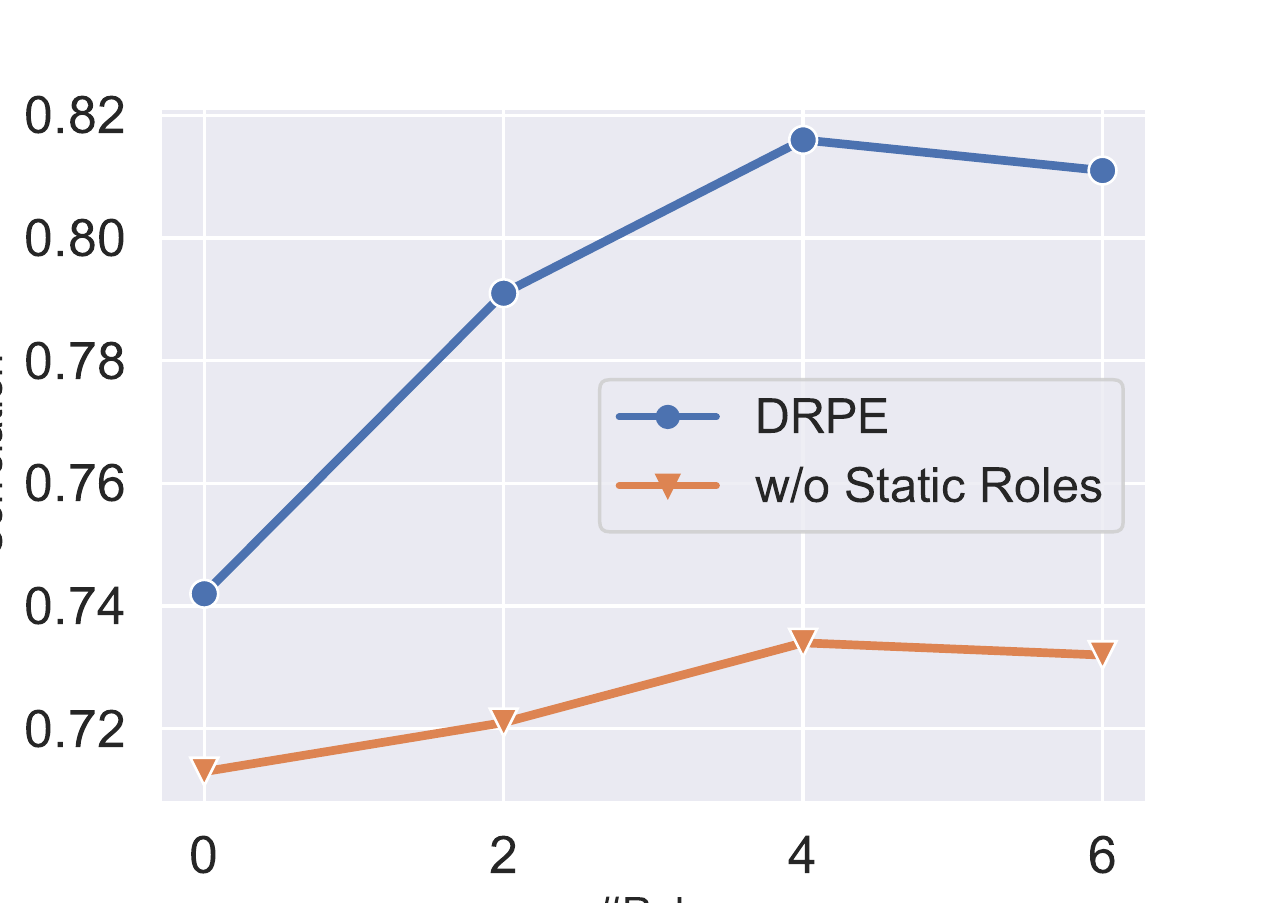}
\end{minipage}\label{fig:case-study-a}
}%
\subfigure[Roles number k w.r.t correlation on BBC2022 dataset.]{
\begin{minipage}[t]{0.4\linewidth}
\centering
\includegraphics[width=0.95\linewidth]{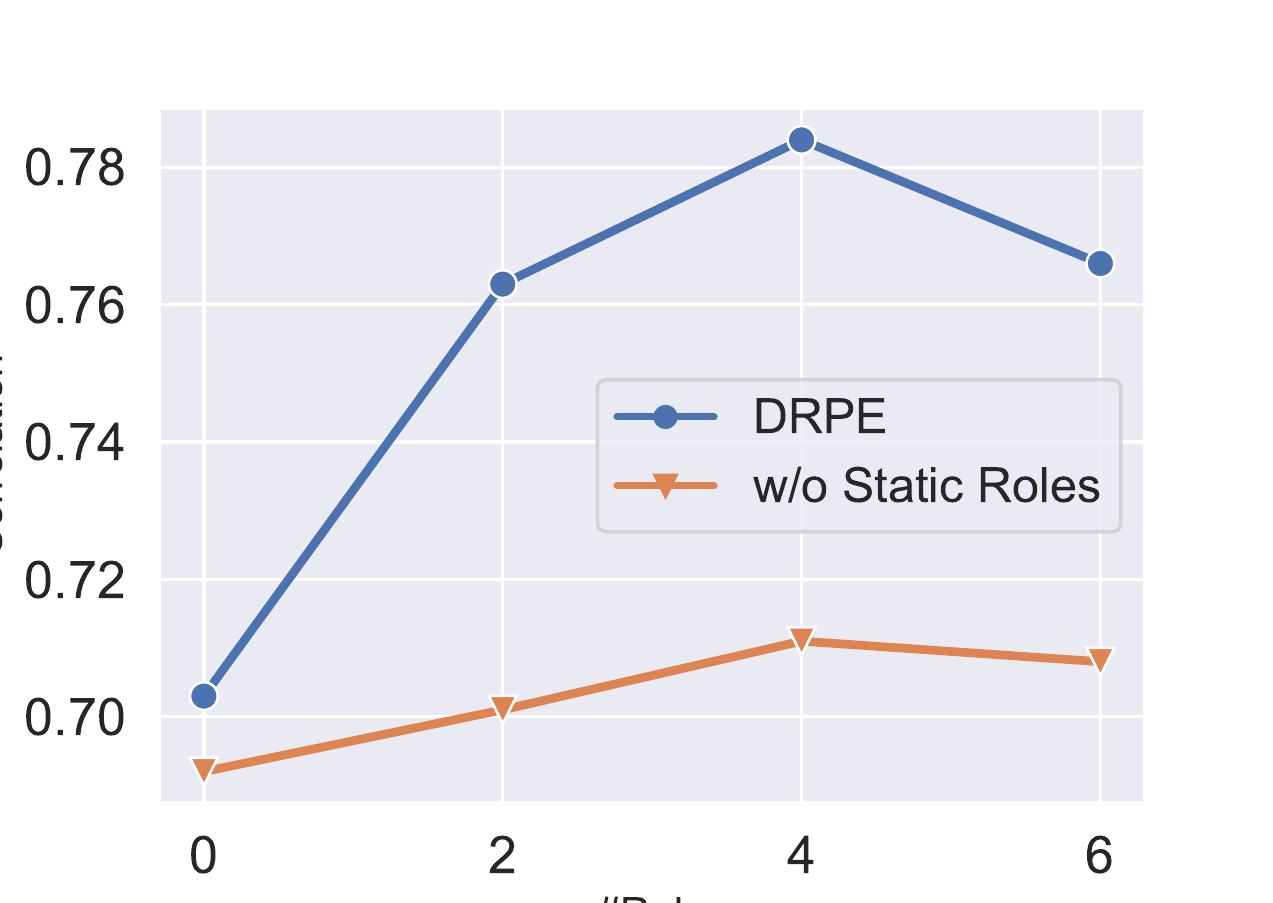}
\end{minipage}\label{fig:case-study-b}
}%
\centering
\caption{Effect of role number on model performance. } \label{fig:retrieval_number}
\end{figure}

\subsubsection{Effects of Hyperparameters}

We test DRPEScore and \uline{w/o Static Roles} with different subjective role numbers $k$ in [0, 2, 4, 6] on two datasets. Figure~\ref{fig:retrieval_number} shows the experimental results. When $k$ increases from 0 to 6, the experimental indicators first increase and then slightly decrease on the CNN2022 dataset,  when $k$=4, the correlation to human evaluation achieves a peak. On the BBC2022 dataset, experimental results are similar, and more roles~(6) don't bring improvements compared to fewer roles~(4). 


\begin{figure*}[!ht]
    \centering
    \includegraphics[width=0.85\linewidth]{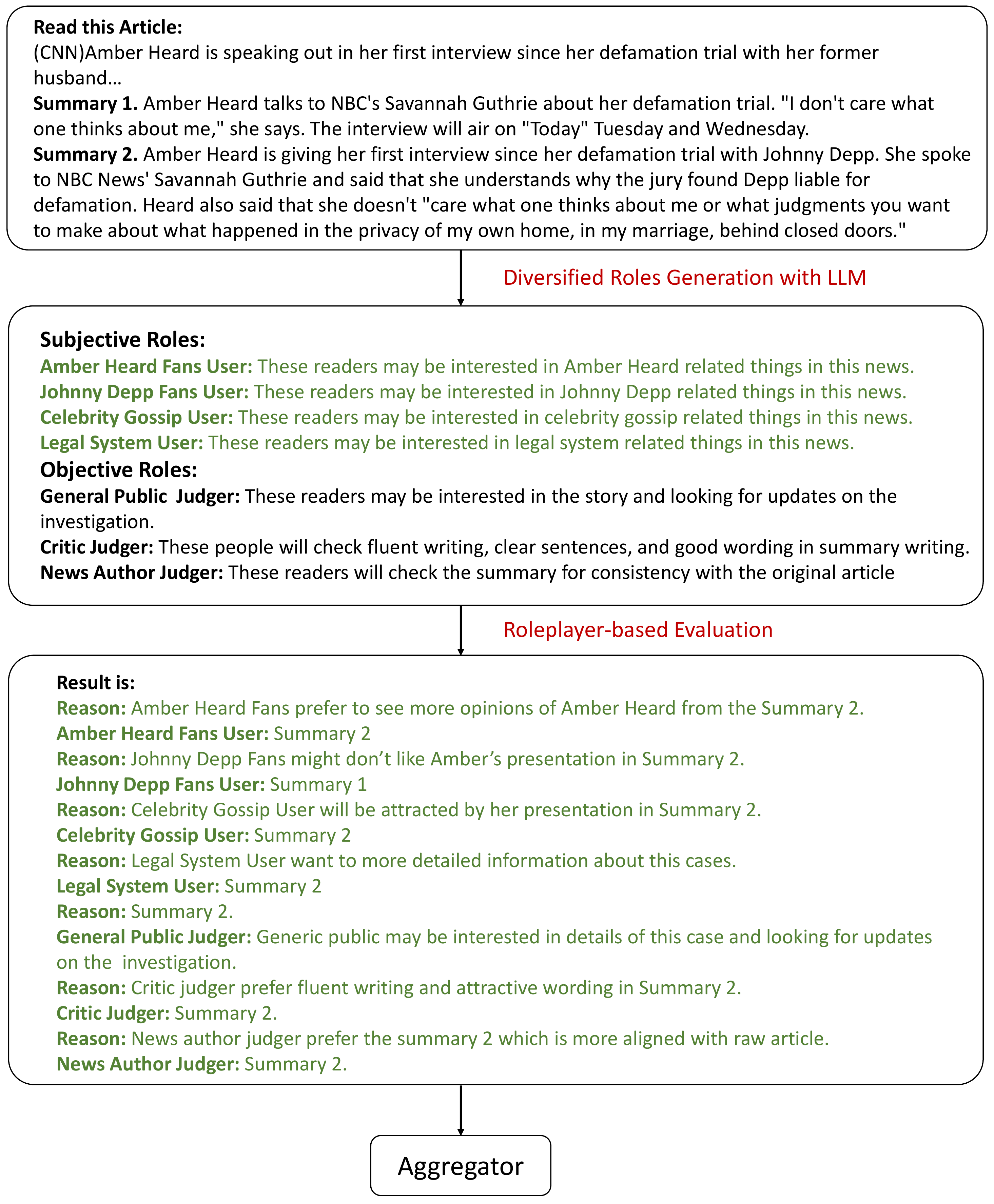}
    \caption{Evaluation procedure for two summaries given a piece of news. We use the green font to represent the content generated by the model. With suitable prompting, the LLM is able to generate relevant roles and generate judgment for these roles.  }
    \label{fig-case}
\end{figure*}

\subsubsection{Qualitative Analysis}
We have previously demonstrated the effectiveness of our model on two summarization tasks. In this section, we conduct a qualitative analysis to explain why DRPEScore can achieve good performance. Figure~\ref{fig-case} shows an example of our model’s evaluation process. Given a news article about the legal dispute between Amber Heard and Johnny Depp, our model has to select a better summary from two candidates. First, we generate several subjective roles based on the news content, such as \emph{Amber Heard Fans User}, \emph{Johnny Depp Fans User}, \emph{Celebrity Gossip User}, and \emph{Legal System User}. These roles are representative of different perspectives and preferences that can be captured by LLMs. Second, we use LLMs to simulate each user and judger and compare the two summaries. We employ a batch prompting mechanism to accelerate the inference procedure. Notably, LLMs predict that the \emph{Johnny Depp Fans User}, who might have a negative attitude towards \emph{Amber Heard Fans User}, will favor summary 1.


\section{Conclusion}
We propose DRPE, a new comparison-based method for evaluating generated text against gold standard references. Our DRPE is designed to be simple and task-agnostic. Our experiments illustrate that DRPE could provide a human-like ability to conduct a comprehensive evaluation, especially
on challenging long text generation like summarization tasks.  In future work, we look forward to exploring the capabilities of LLMs as judgers on more text evaluation tasks and reducing the computation cost.

%
%
%
%

\bibliography{anthology,custom}

\begin{thebibliography}{10}
\providecommand{\url}[1]{\texttt{#1}}
\providecommand{\urlprefix}{URL }
\providecommand{\doi}[1]{https://doi.org/#1}

\bibitem{banerjee2005meteor}
Banerjee, S., Lavie, A.: Meteor: An automatic metric for mt evaluation with
  improved correlation with human judgments. In: Proceedings of the acl
  workshop on intrinsic and extrinsic evaluation measures for machine
  translation and/or summarization. pp. 65--72 (2005)

\bibitem{brown2020language}
Brown, T., Mann, B., Ryder, N., Subbiah, M., Kaplan, J.D., Dhariwal, P.,
  Neelakantan, A., Shyam, P., Sastry, G., Askell, A., et~al.: Language models
  are few-shot learners. Advances in neural information processing systems
  \textbf{33},  1877--1901 (2020)

\bibitem{fabbri2021summeval}
Fabbri, A.R., Kry{\'s}ci{\'n}ski, W., McCann, B., Xiong, C., Socher, R., Radev,
  D.: Summeval: Re-evaluating summarization evaluation. TACL  \textbf{9},
  391--409 (2021)

\bibitem{gao2014modeling}
Gao, J., Pantel, P., Gamon, M., He, X., Deng, L.: Modeling interestingness with
  deep neural networks. In: Proceedings of the 2014 conference on empirical
  methods in natural language processing (EMNLP). pp. 2--13 (2014)

\bibitem{goyal2022news}
Goyal, T., Li, J.J., Durrett, G.: News summarization and evaluation in the era
  of gpt-3. arXiv preprint arXiv:2209.12356  (2022)

\bibitem{hermann2015teaching}
Hermann, K.M., Kocisky, T., Grefenstette, E., Espeholt, L., Kay, W., Suleyman,
  M., Blunsom, P.: Teaching machines to read and comprehend. NIPS  \textbf{28}
  (2015)

\bibitem{hidi1986interestingness}
Hidi, S., Baird, W.: Interestingness—a neglected variable in discourse
  processing. Cognitive science  \textbf{10}(2),  179--194 (1986)

\bibitem{kojima2022large}
Kojima, T., Gu, S.S., Reid, M., Matsuo, Y., Iwasawa, Y.: Large language models
  are zero-shot reasoners. arXiv preprint arXiv:2205.11916  (2022)

\bibitem{lin2004rouge}
Lin, C.Y.: Rouge: A package for automatic evaluation of summaries. In: Text
  summarization branches out. pp. 74--81 (2004)

\bibitem{ma2018results}
Ma, Q., Bojar, O., Graham, Y.: Results of the wmt18 metrics shared task: Both
  characters and embeddings achieve good performance. In: Proceedings of the
  third conference on machine translation: shared task papers. pp. 671--688
  (2018)

\bibitem{nallapati2016abstractive}
Nallapati, R., Zhou, B., Gulcehre, C., Xiang, B., et~al.: Abstractive text
  summarization using sequence-to-sequence rnns and beyond. arXiv preprint
  arXiv:1602.06023  (2016)

\bibitem{narayan2018don}
Narayan, S., Cohen, S.B., Lapata, M.: Don't give me the details, just the
  summary! topic-aware convolutional neural networks for extreme summarization.
  arXiv preprint arXiv:1808.08745  (2018)

\bibitem{papineni2002bleu}
Papineni, K., Roukos, S., Ward, T., Zhu, W.J.: Bleu: a method for automatic
  evaluation of machine translation. In: ACL. pp. 311--318 (2002)

\bibitem{reimers2019sentence}
Reimers, N., Gurevych, I.: Sentence-bert: Sentence embeddings using siamese
  bert-networks. In: EMNLP. pp. 3982--3992 (2019)

\bibitem{rubner2000earth}
Rubner, Y., Tomasi, C., Guibas, L.J.: The earth mover's distance as a metric
  for image retrieval. International journal of computer vision
  \textbf{40}(2), ~99 (2000)

\bibitem{sanh2021multitask}
Sanh, V., Webson, A., Raffel, C., Bach, S.H., Sutawika, L., Alyafeai, Z.,
  Chaffin, A., Stiegler, A., Scao, T.L., Raja, A., et~al.: Multitask prompted
  training enables zero-shot task generalization. arXiv preprint
  arXiv:2110.08207  (2021)

\bibitem{wang2022self}
Wang, X., Wei, J., Schuurmans, D., Le, Q., Chi, E., Zhou, D.: Self-consistency
  improves chain of thought reasoning in language models. arXiv preprint
  arXiv:2203.11171  (2022)

\bibitem{wei2022chain}
Wei, J., Wang, X., Schuurmans, D., Bosma, M., Chi, E., Le, Q., Zhou, D.: Chain
  of thought prompting elicits reasoning in large language models. arXiv
  preprint arXiv:2201.11903  (2022)

\bibitem{yuan2022selecting}
Yuan, X., Wang, T., Wang, Y.H., Fine, E., Abdelghani, R., Lucas, P.,
  Sauz{\'e}on, H., Oudeyer, P.Y.: Selecting better samples from pre-trained
  llms: A case study on question generation. arXiv preprint arXiv:2209.11000
  (2022)

\bibitem{zhang2019bertscore}
Zhang, T., Kishore, V., Wu, F., Weinberger, K.Q., Artzi, Y.: Bertscore:
  Evaluating text generation with bert. arXiv preprint arXiv:1904.09675  (2019)

\bibitem{zhao2019moverscore}
Zhao, W., Peyrard, M., Liu, F., Gao, Y., Meyer, C.M., Eger, S.: Moverscore:
  Text generation evaluating with contextualized embeddings and earth mover
  distance. arXiv preprint arXiv:1909.02622  (2019)

\bibitem{zhou2022least}
Zhou, D., Sch{\"a}rli, N., Hou, L., Wei, J., Scales, N., Wang, X., Schuurmans,
  D., Bousquet, O., Le, Q., Chi, E.: Least-to-most prompting enables complex
  reasoning in large language models. arXiv preprint arXiv:2205.10625  (2022)

\end{thebibliography}

\bibliographystyle{splncs04}
\end{document}